\documentclass[runningheads,a4paper]{llncs}
\usepackage{amssymb,amsmath}
\usepackage{graphicx}

\usepackage[utf8]{inputenc}

\usepackage[numbers, sectionbib]{natbib}
\usepackage[ruled]{algorithm2e}
\usepackage{url}

\usepackage[usenames,dvipsnames]{color}
\usepackage[colorlinks=true]{hyperref}
\hypersetup{
  colorlinks,
 citecolor=Blue,
 linkcolor=Red,
 urlcolor=Violet}

\newcommand{\keywords}[1]{\par\addvspace\baselineskip
\noindent\keywordname\enspace\ignorespaces#1}

\newcommand{\bx}{\bold{x}}
\newcommand{\bX}{\bold{X}}

\newcommand{\bmu}{\boldsymbol{\mu}}

\begin{document}

\mainmatter  

\title{Simultaneous Clustering and Model Selection for Multinomial Distribution: A Comparative Study}

\titlerunning{Simultaneous Clustering and Model Selection for Multinomial Distribution}

%
\author{Md. Abul Hasnat%
\and Julien Velcin\and St\'ephane Bonnevay \and Julien Jacques}
\authorrunning{Hasnat, Velcin, Bonnevay and Jacques}

\institute{Laboratoire ERIC, Universit\'e Lumi\`ere Lyon 2, Lyon, France\\
md.hasnat@univ-lyon2.fr, julien.velcin@univ-lyon2.fr, stephane.bonnevay@univ-lyon1.fr, julien.jacques@univ-lyon2.fr}

%
%

\toctitle{Lecture Notes in Computer Science}
\tocauthor{Authors' Instructions}
\maketitle

\begin{abstract}
In this paper, we study different discrete data clustering methods, which use the Model-Based Clustering (MBC) framework with the Multinomial distribution. Our study comprises several relevant issues, such as initialization, model estimation and model selection. Additionally, we propose a novel MBC method by efficiently combining the partitional and hierarchical clustering techniques. We conduct experiments on both synthetic and real data and evaluate the methods using accuracy, stability and computation time. Our study identifies appropriate strategies to be used for discrete data analysis with the MBC methods. Moreover, our proposed method is very competitive w.r.t. clustering accuracy and better w.r.t. stability and computation time.
\end{abstract}

\section{Introduction}
\label{sec:intro}
Model-Based Clustering (MBC) estimates the parameters of a statistical model for the data and produces probabilistic clustering \cite{fraley2002model, fraley2007model, melnykov2010finite, zhong2003unified}. To use the MBC method for clustering data as well as automatically selecting $K$ (number of clusters), it is necessary to generate a set of candidate models. 
A simple approach to generate these models is to separately estimate them using an Expectation-Maximization (EM) method \cite{mclachlan2008em} with $K=1,\ldots,K_{max}$.
However, it can be computationally inefficient for higher dimensional data and higher $K_{max}$ value. 
%

Figueiredo and Jain \cite{figueiredo2002unsupervised} proposed a MBC method that integrates both model estimation and  selection task within a single EM algorithm.
A different strategy, called hybrid MBC \cite{zhong2003unified}, generates a hierarchy of models from $K_{max}$ clusters by merging the parameters. Indeed, such an approach naturally saves computation time as it does not explicitly learn  $K=K_{max}-1, \ldots 1$ components models from the data. In this paper, we propose a hybrid MBC method with the Multinomial Mixture (MM) model and then empirically compare it with other MBC methods. Moreover, we explicitly addresses two related issues: (1) initialization \cite{biernacki2003choosing}: how to set the initial parameters for the EM method and (2) model selection \cite{biernacki2000assessing}: which criterion to use for selecting the best model. Therefore, based on an empirical study, we aim to answer the following questions: (a) which method should be used for initialization? (b) how to efficiently generate a set of models? (c) what is the difference among ``learning from data'' and ``estimating from $ K_{max}$ model  parameters''? and (d) what is the best model selection method?

Our overall contribution is to perform a comparative study among different MBC methods with the MM. Individually, we: (1) propose (Sec. \ref{ssec:mbhc_mm}) a novel MBC method and compare it with the state-of-the-art methods; (2) perform empirical study on different initialization methods (Sec. \ref{ssec:init_em}) and (3) compare different model selection methods (Sec. \ref{ssec:model_selection}). We conduct experiments with synthetic and real text data (for document clustering \cite{zhong2003unified}) and identify particular methods that should be used for initialization, candidate models estimation and model selection. Therefore, the above contributions and experiments will naturally answer the questions raised at the end of the previous paragraph.

In the remaining part of this paper, we study the background and related work in Sec. \ref{sec:rel_work}, discuss different methods in Sec. \ref{sec:methodologies}, present the experimental results with discussion in Sec. \ref{sec:exp_res} and finally draw conclusions in Sec. \ref{sec:conclusions}.
%
%
\section{Background and Related Work}
\label{sec:rel_work}
%
%
Model-Based Clustering (MBC) \cite{fraley2002model, melnykov2010finite} is a well-established method for cluster analysis and unsupervised learning. 
MBC assumes a probabilistic model (e.g., mixture model) for the data and then estimates the model parameters by optimizing an objective function (e.g., model likelihood). The Expectation Maximization (EM) \cite{mclachlan2008em} is mostly used in MBC to estimate the model parameters. EM consists of an Expectation step (E-step) and a Maximization step (M-step) which are iteratively employed to maximize the log likelihood of the data. 

MBC methods have been exploited with the Gaussian distribution to analyze continuous data \cite{fraley2002model, melnykov2010finite, figueiredo2002unsupervised, biernacki2000assessing, fraley2007model}. Besides, they have been proposed to analyze discrete data using the Multinomial distribution \cite{meilua2001experimental, silvestre2014identifying} and directional data using the directional distributions \cite{banerjee2005aClustering, hasnat2014wmm, hasnat2015mbhcfmm}.
In this paper, we only study and compare the MBC methods with the Multinomial distribution.

The Multinomial Mixture (MM) is a statistical model which has been used for cluster analysis with discrete data \cite{meilua2001experimental, zhong2005generative, silvestre2014identifying}. Meil{\u{a}} and Heckerman \cite{meilua2001experimental} studied the MBC methods with MM and compared them w.r.t. accuracy, time and number of clusters. They found that the EM method significantly outperforms others, which motivates us to solely focus on the EM related approaches. 

Initialization of the EM method has significant impact on the clustering results \cite{mclachlan2008em, biernacki2003choosing, maitra2009initializing}, because with different initializations it may converge to different values of the likelihood function, some of which can be local maxima, i.e., sub-optimal results. To overcome this, several initialization strategies have been proposed, see \cite{biernacki2003choosing} for details. Meil{\u{a}} and Heckerman \cite{meilua2001experimental}  investigated three initialization strategies for the EM with MM. In this paper, we consider their \cite{meilua2001experimental} observations as well as empirically evaluate additional initialization methods for the EM method which were discussed by Biernacki et al. \cite{biernacki2003choosing}. 

In order to automatically select $K$ (number of components), MBC method can be used by first generating a set of candidate models with different values of $K$ and then selecting the optimal model using a model selection criterion \citep{fraley2002model, melnykov2010finite}. This strategy needs to address two issues: (a) how to generate the models? and (b) how to select the best model? This paper considers both of these issues. Particularly, we focus on the candidate models generation task and propose a novel solution based on the Hybrid MBC (HMBC) \cite{zhong2003unified} method.

HMBC method is a two-staged model that exploits both partitional and hierarchical clustering. It begins with a partitional clustering with $K_{max}$ clusters and then use the Hierarchical Agglomerative Clustering (HAC) on those cluster parameters to generate a hierarchy of mixture models. It has differences with the Model-Based Hierarchical Clustering (MBHC) which employs the HAC on each data point \cite{fraley2002model}. In practice, for a large number of samples, such MBHC method is inefficient w.r.t. the required time and memory \cite{zhong2003unified}. Several HMBC methods have been proposed with different probability distributions, see \cite{zhong2003unified}, \cite{garcia2010simplification}, \cite{hasnat2015mbhcfmm} and \cite{vaithyanathan2000model}. Among these, \cite{vaithyanathan2000model} proposed a method in the context of Bayesian analysis. However, it requires an explicit analysis of the features, which can be computationally inefficient for higher dimensional data. An efficient mixture model simplification/fusion method is recently proposed in \cite{garcia2010simplification} for the Gaussian distribution and in \cite{hasnat2015mbhcfmm, hasnat2014wmm} for the directional distributions. They use information divergences among the mixture models. In this paper, we follow a similar approach and  propose a novel HMBC method with the MM.

Model selection is one of the most prominent issues in cluster analysis \cite{melnykov2010finite, figueiredo2002unsupervised,  biernacki2000assessing,  fraley2007model}. In general, a statistical model selection criterion is often used with the MBC method, which is also called the parsimony-based approach \citep{melnykov2010finite}. See \cite{figueiredo2002unsupervised} for a list of different criteria. A different approach performs model selection by analyzing an evaluation graph, see \cite{salvador2004determining} for such a method called the L-method. To select model with MM, \cite{meilua2001experimental} uses the likelihood value. Recently, \cite{silvestre2014identifying} proposed the Minimum Message Length (MML) criterion for the MM. 
In this paper, we aim to present a comparative study among these methods.

This paper has similarity with two previous work \cite{meilua2001experimental} and \cite{silvestre2014identifying}. However, the key differences are: (1) it proposes a novel method to efficiently generate candidate models; (2) investigate additional initialization methods proposed in \cite{biernacki2003choosing} and (3) explore a wide range of model selection methods.
\section{Methodologies}
\label{sec:methodologies}
In the following sub-sections, first we present the model for the data, then discuss the relevant algorithms and finally propose a complete clustering method.
\subsection{Multinomial Mixture Model}
Let $\bx_i= x_{i,1}, x_{i,2}, \ldots, x_{i,D}$ is a $D$ dimensional discrete count vector of order $V$, i.e. $\sum_{d=1}^{D}x_{i,d}=V$. Moreover, $\bx_i$ is assumed to be an independent realization of the random variable $\bX$, which follows a $V$-order Multinomial distribution \cite{bishop2006pattern}: 
\begin{equation}
\label{eq:mult_dist}
\mathcal{M}(\bx_i |V,  \bmu) = \begin{pmatrix}
 V \\ 
 x_{i,1}, x_{i,2}, \ldots, x_{i,D} 
\end{pmatrix} \prod_{d=1}^{D} \mu_d^{x_{i,d}}
\end{equation}
here, $\bmu$ is the $D$ dimensional parameter with $0 \leq \mu_d \leq 1$ and $\sum_{d=1}^{D}\mu_d=1$. The set of samples can be modeled with a Multinomial Mixture (MM) model of $K$ components:
\begin{equation}
\label{eq:mm}
f\left ( \bx_i|\Theta_{K} \right ) = \sum_{k=1}^{K}\pi_{k} \, \mathcal{M}(\bx_i | V,  \bmu_k) 
\end{equation}

In Eq. (\ref{eq:mm}), $\Theta_{K} = \left \{ (\pi_1,\bmu_1), \ldots , (\pi_K,\bmu_K)\right \}$ is the set of model parameters, $\pi_k$ is the mixing proportion with $\sum_{k=1}^{K}\pi_{k}=1$ and $\mathcal{M}(\bx_i | V, \bmu_k)$ is the density function (Eq. (\ref{eq:mult_dist})) associated with the $k^{th}$ cluster.
\subsection{Expectation Maximization Method}
\label{ssec:est_mm_parameters}
To cluster data with the model (Eq. (\ref{eq:mm})), we estimate its parameters using an Expectation Maximization (EM) \cite{mclachlan2008em} method that maximizes the log-likelihood:
\begin{equation}
\label{eq:llh_comp}
L\left ( \Theta \right ) = \sum_{i=1}^{N} log \sum_{k=1}^{K} \pi_k \mathcal{M}\left ( \bx_i|\bmu_k \right )
\end{equation}
where $N$ is the number of samples. In the Expectation step (E-step), we compute posterior probability as:
\begin{equation}
\label{eq:posterior_cl}
\rho_{i,k} = p\left (z_i=k|\mathbf{x}_i \right ) = \frac{\pi_k \, \prod_{d=1}^{D}\mu_{k,d}^{x_{i,d}}}
{\sum_{l=1}^{K}\pi_{l} \, \prod_{d=1}^{D}\mu_{l,d}^{x_{i,d}}}
\end{equation}
where $z_i \in \lbrace 0,1\rbrace ^ K$ denotes the cluster label of the $i^{th}$ sample. In the Maximization step (M-step), we update $\pi_k$ and $\mu_{k,d}$ as:
\begin{equation}
\label{eq:maximization_cl}
\pi_k=\frac{1}{N}\sum_{i=1}^{N}\rho_{i,k}
\;\;\;
\text{and}
\;\;\; 
\mu_{k, d}=\frac{\sum_{i=1}^{N}\rho_{i,k} \,x_{i, d}}{\sum_{i=1}^{N} \sum_{r=1}^{D} \rho_{i,k} \,x_{i, r}}
\end{equation}
The E and M steps run iteratively until certain convergence criterion (e.g., difference of log-likelihood) is met or until a maximum number of iterations. 

\subsection{Initialization for the EM Method}
\label{ssec:init_em}
The EM method requires the initial values of the parameters as an input. We examine the following five methods to initialize the EM:
\begin{itemize}
\item \textbf{Random:} set the initial values randomly with $0 \leq \mu_d \leq 1$ and $\sum_{d=1}^{D}\mu_d=1$.
\item \textbf{rndEM \cite{maitra2009initializing}:} run a large number of random start and select the one which provides maximum likelihood value (Eq. (\ref{eq:llh_comp})). 
\item \textbf{Small EM (smEM) \cite{biernacki2003choosing}:} run multiple short runs of randomly initialized EM and choose the one with the maximum likelihood value. Here, short run means we do not wait until convergence and stop the algorithm when limited number of EM iterations is completed.
\item \textbf{Classification EM (CEM) \cite{biernacki2003choosing}:} it is similar to the smEM, except a classification stage is inserted between the E and M steps. The classification step involves assigning each point to one of the $K$ components using the conditional probabilities (Eq. (\ref{eq:posterior_cl})) computed in the E step.
\item \textbf{Stochastic EM (SEM) \cite{biernacki2003choosing}:} it is similar to the smEM, except a stochastic step is inserted between the E and M steps. The stochastic step assigns $\bx_i$ at random to one of the mixture components $K$ according to the Multinomial distribution with the conditional probabilities  (Eq. (\ref{eq:posterior_cl})).
\end{itemize}
%
%
%
\subsection{Candidate Models Generation}
\label{ssec:mm_model_generation}

\subsubsection{Multiple EM (Mul-EM):}
\label{sssec:mul_EM}
This is the simplest way to generate the candidate models. In this approach, the EM method is run $K_{max}$ times to generate the candidate models with $K = {1,\ldots,K_{max}}$ clusters.

\subsubsection{Integrated-EM (Int-EM):}
\label{sssec:int_EM}
This approach  \cite{ figueiredo2002unsupervised, silvestre2014identifying} do not explicitly generates the candidate models. Instead, it employs a single EM method that estimates the MM with $K$ clusters and evaluate it at the same time. It begins with $K=K_{max}$ clusters and estimate its parameter. Then it annihilates a cluster with minimum $\pi_k$ and estimate parameters with $K-1$ clusters. This process continues within a single EM method until $K=1$. See the EM-MML algorithm of \cite{silvestre2014identifying} for details.

\subsubsection{EM followed by Hierarchical Agglomerative Clustering (EM-HAC):}
\label{sssec:hac_model_generation}
This is our proposed model generation method, which aim is to generate a hierarchy of Multinomial Mixture (MM) models. Therefore, we exploit the Hierarchical Agglomerative Clustering (HAC) on the mixture model parameters $\hat{\Theta}_K$. In general, the HAC permits a variety of choices based on three principal issues: (a) the dissimilarity measure between clusters; (b) the criterion to select the clusters to be merged and (c) the representation of the merged cluster.

We use the symmetric Kullback–Leibler Divergence \cite{bishop2006pattern} (sKLD) as a measure of the dissimilarity between two Multinomial distributions as:
\begin{equation}
\label{eq:kld}
sKLD = \frac{D_{KL}\left ( \bmu_a, \bmu_b \right ) + D_{KL}\left ( \bmu_b, \bmu_a \right )}{2} \,, \text{where,} \,
D_{KL}\left ( \bmu_a, \bmu_b \right ) = \sum_{d=1}^{D} \mu_{a,d}\, ln\left ( \frac{\mu_{a,d}}{\mu_{b,d}} \right )
\end{equation}
We choose ``minimum sKLD'' as the merging criterion (issue (b)).
Besides we use the ``complete linkage'' criteria which is determined empirically.

In this clustering strategy, the set of models is represented by their parameters. After determining the clusters to be merged, similar to \cite{garcia2010simplification, hasnat2015mbhcfmm}, we compute
the merged cluster parameters (issue (c)) as:
\begin{equation}
\label{eq:clrep}
\pi_{merged} = \sum_{l \in \hat{\Theta}_{sub}}\pi_{l,k} \,\,\, \text{and} \,\,\, \bmu_{merged} = \frac{\sum_{l \in \hat{\Theta}_{sub}}\pi_{l}\bmu_l}{\pi_{merged}}
\end{equation}
%
where $\hat{\Theta}_{sub} \subseteq \hat{\Theta}_{K_{max}}$. As an outcome, we obtain a set of MMs with different $K$, which will be explored further for model selection.
%
\subsection{Model Selection}
\label{ssec:model_selection}
Consider that, after HAC we have a set of MMs with $K_{max},\ldots,1$ components. The task of model selection can be defined as selecting the mixture model with $K_{o}$ components such that $\hat{\Theta}_{K_o} = \left \{ (\hat{\pi}_1, \hat{\mu}_1),\ldots,(\hat{\pi}_{K_o}, \hat{\mu}_{K_o})\right \}$. We consider parsimony-based \cite{melnykov2010finite} and evaluation graph based \cite{salvador2004determining} methods in this work.

In the parsimony-based method\cite{melnykov2010finite}, an objective function is employed, which minimizes certain model selection criteria. Such criteria involve the negative log likelihood augmented by a penalizing function in order to take into account the complexity of the model. One of the most widely used criteria is called the Bayesian Information Criterion (BIC) \cite{fraley2002model}:
\begin{equation}
\label{eq:bic}
BIC(K) = -2 L(\hat{\Theta}) + \nu log \left ( N \right )
\end{equation}
where $\nu=KD-1$ is the number of free parameters of the MM. 
The Integrated Completed Likelihood (ICL) criterion adds BIC with the mean entropy \citep{biernacki2000assessing}:
\begin{equation}
\label{eq:icl_compute}
ICL(K)= BIC(K) - 2\sum_{i=1}^{N}\text{log}\left ( p(z_{i}|\bx_{i}) \right )
\end{equation}
where $p(z_{i}|\bx_{i})$ is the conditional probability of the classified class label $z_{i} \in \{1,\ldots,K\}$ for the sample $\bx_{i}$. The Minimum Message Length (MML) criterion, which has been recently proposed for MM, has the following form \cite{silvestre2014identifying}: 
\begin{equation}
\label{eq:mml_compute}
MML(K) = \frac{D}{2} \sum_{k:\hat{\pi}_k> 0} \text{log}\left ( \frac{N \, \hat{\pi}_k}{12} \right ) + \frac{K_{nz}}{2} \text{log}\frac{N}{12} + \frac{K_{nz}\left ( D+1 \right )}{2} - L(\hat{\Theta})
\end{equation}
where $K_{nz}$ is the number of clusters with non-zero probabilities.
After computing the values of the model selection criteria for different $K\in \{ 1,...,K_{max}\}$, we select $K_o$ as the one that provides the minimum value of certain criterion.

For the evaluation graph based method, we consider the L-method (see \citep{salvador2004determining} for details), where the knee point is detected in the plot constructed from the BIC values. The idea is to fit two lines at the left and right side of each point within the range 2,...,$K_{max}-1$. Finally, select the point as $K_{o}$ that minimizes the total weighted root mean squared error.
\subsection{Complete clustering method with MM}
\label{ssec:mbhc_mm}
We propose a complele clustering method with the MM which clusters data and selects the number of clusters automatically. It consists of the following steps:
\begin{itemize}
\item \textbf{\textit{Step 1:}} Apply the EM algorithm (Sec. \ref{ssec:est_mm_parameters}) to estimate MM parameters with $K_{max}$ clusters, i.e., $\hat{\Theta}_{k_{max}}$.
\item \textbf{\textit{Step 2:}} Apply the HAC method (Sec. \ref{sssec:hac_model_generation}) on $\hat{\Theta}_{k_{max}}$ to generate a set of models $\{\hat{\Theta}_k \}_{k=k_{max}-1,...,2}$.
\item \textbf{\textit{Step 3:}} Apply a model selection method (Sec. \ref{ssec:model_selection}) to select $\hat{\Theta}_{K_o}$, i.e., the mixture model with the optimal number of components $K_o$.
\end{itemize}
\section{Experimental Results and Discussion}
\label{sec:exp_res}
We conduct experiments using both simulated and real data. For the evaluation, we compute the Adjusted Rand Index (ARI) \cite{hubert1985comparing}, which is a pair counting based similarity measure among two clustering. Therefore, high value of ARI indicates highly similar clustering and hence high accuracy. For a dataset, we compute the ARI among the clustering result of a particular method and the true labels. 

We evaluate the methods using the clustering accuracy, stability and computation time.  We run each experiment 10 times and record the average value of the ARI as the accuracy, standard deviation of the ARI as the stability\footnote{Stability provides a measure of robustness w.r.t. different initializations. A stable method should provide similar results for different runs, irrespective of its initialization. Therefore, a smaller value of the standard deviation indicates similar results for different runs and hence higher stability of the clustering method.} and the average computation time. 
\subsection{Experimental Datasets}
\label{ssec:exp_datasets}
\subsubsection{Simulated Datasets:}
\label{sssec:simulated_samples}
We draw a finite set of discrete count vectors $\chi= \{\bx_{i}\}_{i,\ldots,N}$ from MMs with different numbers (3, 5 and 10) and types: well-separated (\textbf{\textit{ws}}) and not well-separated (\textbf{\textit{nws}}) of clusters. Similar to \citep{silvestre2014identifying}, the types are verified using the sKLD\footnote{A lower sKLD value among the cluster parameters indicates well-separated clusters, whereas higher value indicates less separation or a certain amount of overlap. Besides computing the sKLD value, we also verified the separation by observing the Bayes error rate among the clusters.} values. We consider samples of different dimensions: 3, 5, 10, 20 and 40. For each MM, we generate 100 sets of data each having 1000 i.i.d. samples. In the synthetic data generation process, first we contruct a MM model with $K$ clusters. The model parameters ($\bmu_k$) for each cluster is sampled from a Dirichlet distribution. The order ($V_k$) of each cluster is sampled randomly from a certain range between $0.5D$ to $1.5D$. After determining the cluster parameters ($\bmu_k$) and orders ($V_k$) we draw the data samples.
\subsubsection{Real Datasets:}
\label{sssec:real_data_samples}
We consider 8 text datasets used in \cite{zhong2005generative}. They consist of discrete count vectors, extracted from different documents collections. The choice was due to its good representation of different characteristics, such as the number of observations (documents), number of features (terms) and the number of clusters. The chosen datasets are listed in Table \ref{tab:real_dataset}. 
We refer the readers to the Sec. 4.2 of \cite{ zhong2005generative} for additional details about the construction of these datasets.

\begin{table}[h]
\centering
\caption{Document text datasets for real data experiments. \textbf{N} denotes number of samples, \textbf{D} denotes number of features and \textbf{K} denotes the number of clusters. The source of the datasets are - \textbf{NG20}: 20 Newsgroups, \textbf{Classic}: ACM/CISI/CRANFIELD/MEDLINE, \textbf{Ohscal}: OHSUMED, \textbf{K1b}: WebACE, \textbf{Hitech}: SJM-TREC, \textbf{Reviews}: SJM-TREC, \textbf{Sports}: SJM-TREC and \textbf{La12}: LAT-TREC. }
\label{my-label}
\begin{tabular}{|c|c|c|c|c|c|c|c|c|}
\hline
          & {\bf NG20}                                               & {\bf Classic}                                                & {\bf Ohscal} & {\bf K1b} & {\bf Hitech}                                        & {\bf Reviews}                                       & {\bf Sports}                                        & {\bf La12}                                          \\ \hline
{\bf N}   & 19949                                                    & 7094                                                         & 11162        & 2340      & 2310                                                & 4069                                                & 8580                                                & 6279                                                \\ \hline
{\bf D}   & 43586                                                    & 41681                                                        & 11465        & 21839     & 10080                                               & 18483                                               & 14870                                               & 31472                                               \\ \hline
{\bf K}   & 20                                                       & 4                                                            & 10           & 6         & 6                                                   & 5                                                   & 7                                                   & 6                                                   \\ \hline
\end{tabular}
\label{tab:real_dataset}
\end{table}

\subsection{Comparisons}
\label{ssec:comparison}
First we compare the initialization strategies listed in Sec. \ref{ssec:init_em} and consistently use the best one for the rest of the experiments. Afterward, we evaluate the model generation methods discussed in Sec. \ref{ssec:mm_model_generation}. Finally, we evaluate the model selection strategies discussed in Sec. \ref{ssec:mm_model_generation}.
\subsubsection{Initialization Methods:}
\label{sssec:eval_initializations}
\begin{figure}
\centering
\includegraphics[scale=0.125]{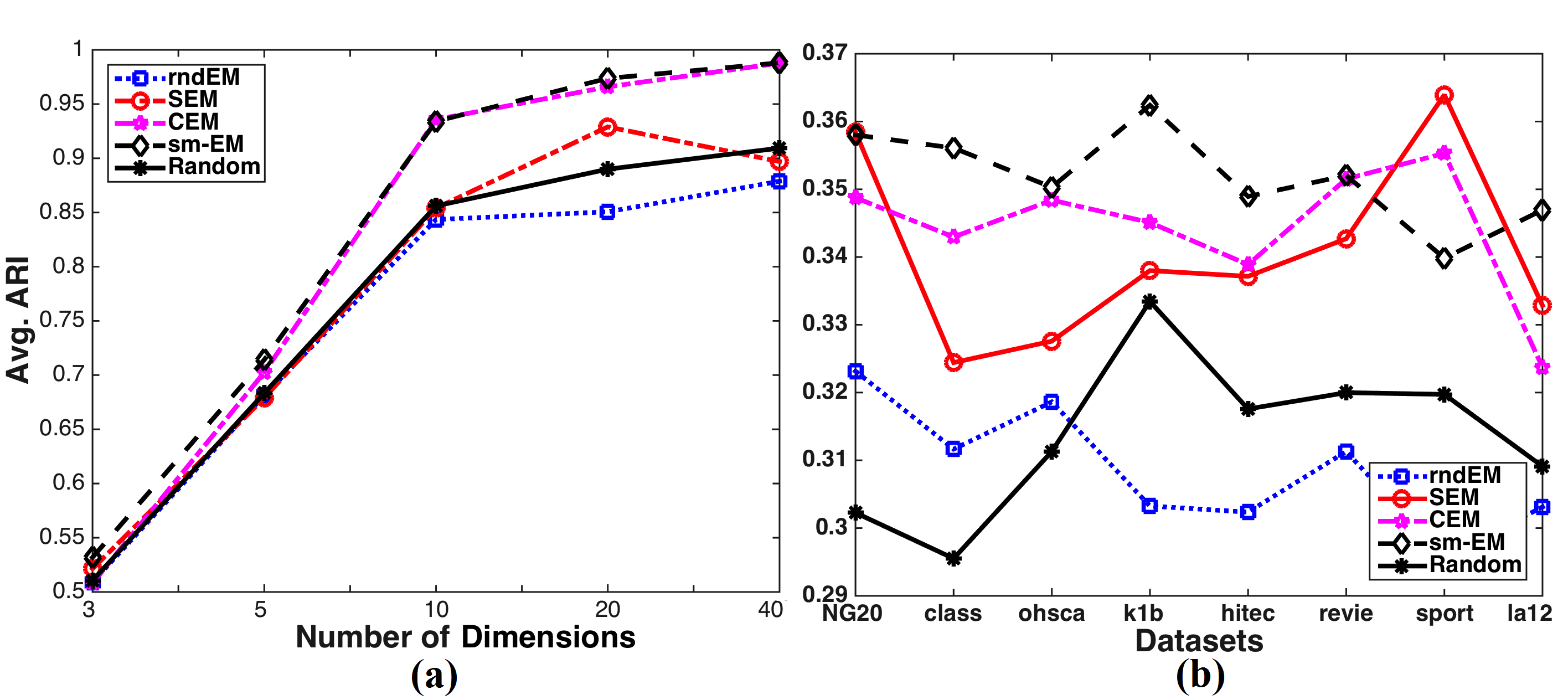} 
\caption{Illustration of the accuracy of the initialization methods, computed from: (a) simulated \textit{\textbf{nws}} samples with $K=3$ and (b) real text datasets.}
\label{fig:init_res}
\end{figure}

The experimental settings for the initialization methods (see Sec. \ref{ssec:init_em}) consist of: 1 trial for \textit{Random}, 100 trials for \textit{rndEM}, 5 trials with 50 maximum EM iterations for \textit{smEM} and \textit{CEM} and 1 trial with 500 maximum EM iterations for \textit{SEM}. The initial parameters obtained from these methods are experimented with the EM method discussed in Sec. \ref{ssec:est_mm_parameters}. Fig. \ref{fig:init_res} illustrates the results w.r.t. the clustering accuracy for both simulated\footnote{Due to limited space, we show results only for \textbf{\textit{nws}} simulated samples with $K=3$.}  and real datasets. From all experimental results we have the following observations:
\begin{itemize}
\item For the simulated data, the \textit{smEM} is the best method while the \textit{CEM} is very competitive. However, for the real data \textit{smEM} provides the best accuracy (except the \textit{sport} dataset). The second choice is the \textit{CEM} method.
\item In terms of stability, \textit{smEM} is the best for simulated data and \textit{CEM} is best for the real data. 
\item In terms of computation time, these methods can be ordered as follows:  $\textit{Random} < \textit{rndEM} < \textit{CEM} < \textit{smEM} < \textit{SEM}$.
\end{itemize}
Similar to \cite{meilua2001experimental}, we emphasize on the clustering accuracy as the main criteria to evaluate the initialization methods. Therefore, we choose the \textit{SEM} method for further experiments. 
\subsubsection{Model Generation Methods:}
\label{sssec:eval_model_generation}
In this experiment, we aim to generate a set of candidate models with the methods discusses in Sec. \ref{ssec:mm_model_generation}. Among them, the \textit{Mul-EM} and \textit{EM-HAC} explicitly generate the models and the \textit{Int-EM} generates them implicitly. All methods are initialized with the \textit{smEM} method. Moreover, same initializations are used in \textit{Int-EM} and \textit{EM-HAC}. Settings of these methods consist of: 100 maximum number of EM iterations, $10^{-5}$ as the convergence threshold for the log-likelihood difference, $K_{min} = 2$ and $K_{max} = 15$, execept for \textit{NG20} $K_{max} = 30$. 
Fig. \ref{fig:mbhc_real_sim_data} illustrates a comparison of these methods w.r.t. the accuracy\footnote{This computation considers that the true numbers of clusters are known.} and stability. Table \ref{tab:comp_time_diff_model_gen} provides a comparison\footnote{Time comparison for the synthetic data provides similar observation as real data. Therefore, to save space we do not present those results.} of the computation time for real data. From all experimental results we have the following observations:
\begin{figure}
\centering
\includegraphics[scale=0.27]{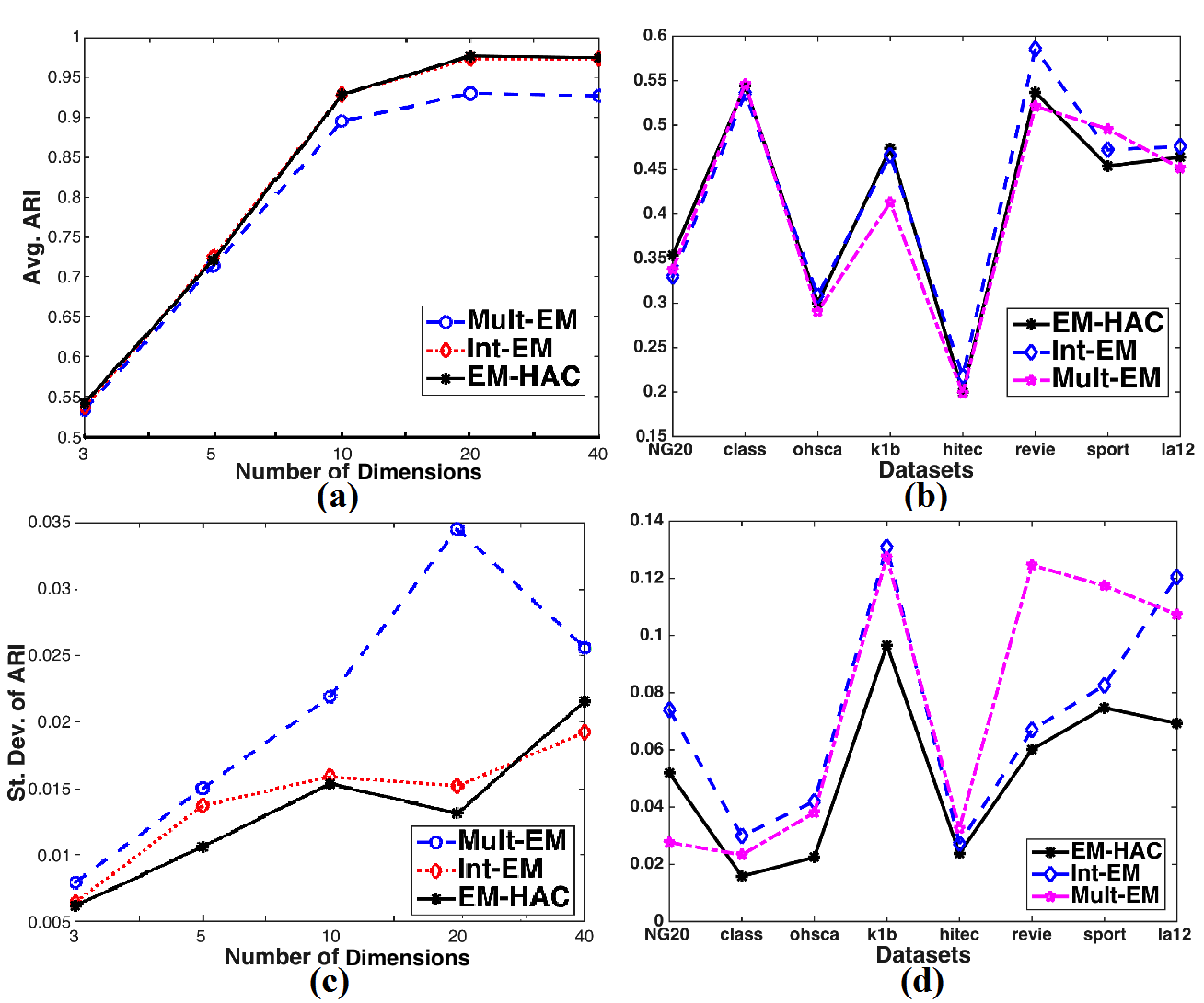} 
\caption{Illustration of the clustering \textit{accuracy} in (a) and (b), and \textit{stability} in (c) and (d) for the model generation methods. (a) and (c) are computed from the simulated \textit{\textbf{nws}} samples; (b) and (d) are computed from real text datasets.}
\label{fig:mbhc_real_sim_data}
\end{figure}
\begin{itemize}
\item For the simulated data: \textit{EM-HAC} and \textit{Int-EM} are very competitive w.r.t. accuracy and time (results not shown). \textit{EM-HAC} is the best on stability. \textit{Mul-EM} was always performing worse except in a very few experiments.
\item For the real\footnote{In this paper we are interested only to compare different MM based MBC methods. We refer readers to \cite{zhong2005generative} for a comparison among different other methods. From \cite{zhong2005generative} we observed that, the \textit{mixmns} (\textit{Mul-EM} in this paper) performs better than the non-MBC methods, such as the \textit{kmns} (k-means) and the \textit{skmns} (spherical k-means).} data, no single method outperforms others w.r.t. the accuracy. \textit{EM-HAC} performs best in 3 datasets, \textit{Int-EM} is best in 4 datasets and \textit{Mul-EM} is best in 1 dataset. \textit{EM-HAC} is best w.r.t. the stability (7 out of 8 datasets). Most interestingly, \textit{EM-HAC} shows significantly better performance in terms of computation time as it is \textbf{$\sim2.5$} times faster than \textit{Int-EM} and \textbf{$\sim9$} times faster than \textit{Mul-EM}.
\end{itemize}
Based on the above experiments and observations, we can suggest that \textit{Int-EM} is preferred when only accuracy is concerned. However, \textit{ EM-HAC} is preferred when stability and time has importantce besides accuracy. 
%
%
\begin{table}[h]
\centering
\caption{Comparison of the computation time (in seconds) among the model generation methods.}
\begin{tabular}{|c|c|c|c|c|c|c|c|c|}
\hline
                 & \textbf{NG20}  & \textbf{Classic} & \textbf{ohscal} & \textbf{k1b} & \textbf{hightech} & \textbf{reviews} & \textbf{sports} & \textbf{la12} \\ \hline
\textbf{EM-HAC}  & \textbf{108.5} & \textbf{6.9}     & \textbf{19.2}   & \textbf{3.8} & \textbf{3.6}      & \textbf{9.9}     & \textbf{17.7}   & \textbf{19.9} \\ \hline
\textbf{Int-EM}  & 353.2          & 10.8              & 42.2            & 9.6          & 8.2               & 21.7             & 46.3            & 44.2          \\ \hline
\textbf{Mult-EM} & 2844.0         & 54.4           & 95.6          & 29.1       & 20.7            & 59.1           & 104.1          & 134.6        \\ \hline
\end{tabular}
\label{tab:comp_time_diff_model_gen}
\end{table} 
\subsubsection{Model Selection Methods:}
\label{sssec:eval_model_generation}
\begin{figure}
\centering
\includegraphics[scale=0.125]{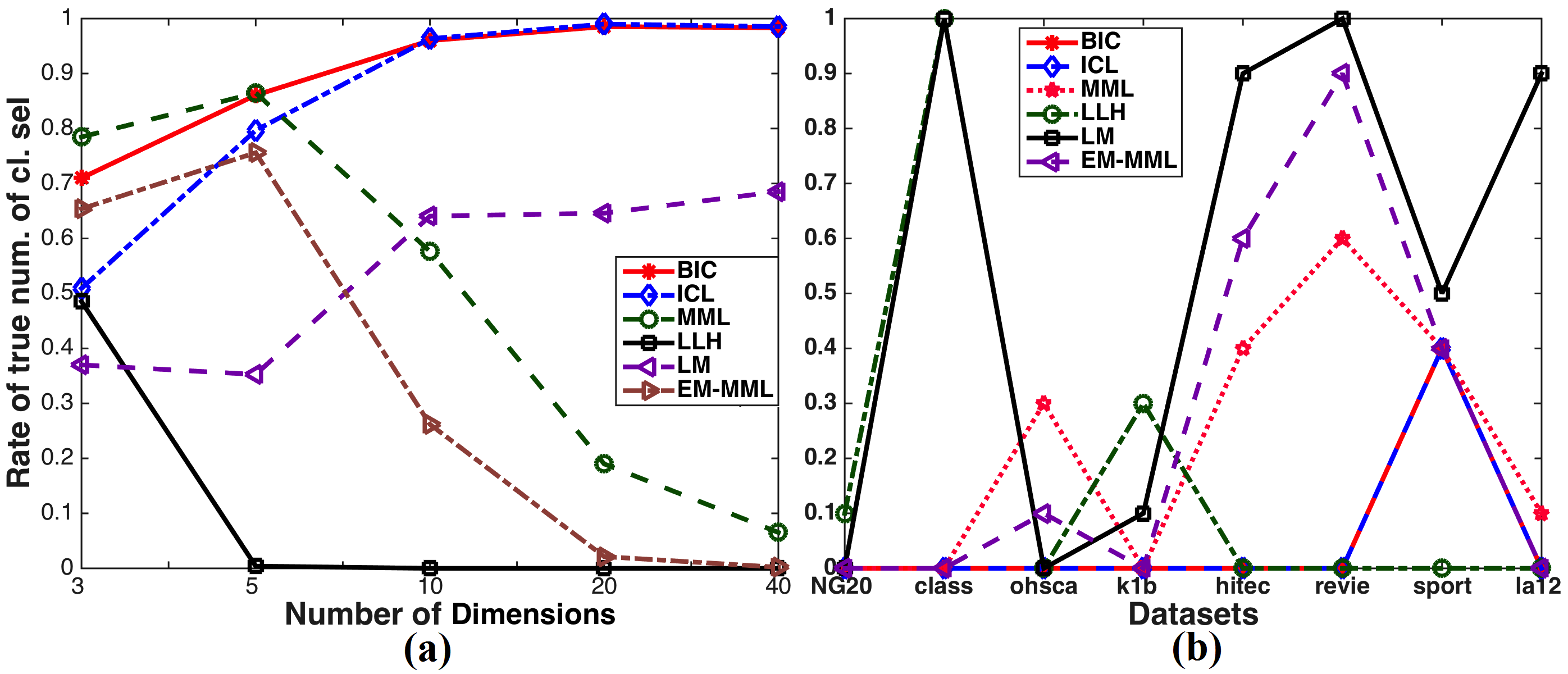} 
\caption{Illustration of the rate of correct model selection, results in (a) are computed from the simulated samples and results in (b) are computed from real text datasets.}
\label{fig:mbhc_comp_sel_data}
\end{figure}

We evaluate different model selection criteria (see Sec. \ref{ssec:model_selection}) with the \textit{EM-HAC}. Moreover, we consider the \textit{MML} with \textit{Int-EM}, also called \textit{EM-MML}, as proposed in \cite{silvestre2014identifying}. Fig. \ref{fig:mbhc_comp_sel_data} illustrates a comparison with both simulated and real data w.r.t. the rate of correct number of components selection. Our observations from these results are as follows:
\begin{itemize}
\item For the simulated data: \textit{BIC} provides the best rate (except $K=3$). \textit{ICL} is equivalent to the \textit{BIC} for higher $K$. Rate of \textit{MML} decreases with the increase of $K$. Moreover, \textit{MML} performs better with \textit{EM-HAC} rather than with \textit{Int-EM}. The \textit{LM} provides mediocre accuracy for all clusters.	The \textit{LLH} criterion fails significantly.
\item For the real data: \textit{LM} provides very good ($\sim90\%$) rate for 4 (\textit{classic, high-tech, review} and \textit{la12}) datasets. Among the other methods, \textit{MML} shows success in the \textit{review} dataset, \textit{LLH} is successful for the \textit{classic} dataset.
\end{itemize}
From the above observations we realize that, the L-method (\textit{LM}) is the best choice with the proposed clustering method. However, we want to emphasize that it is yet necessary to conduct further research on the model selection issue as there is no single method which uniquely provides reasonable rate for all data.
\section{Conclusions}
\label{sec:conclusions}
In this paper, we present a comparative study among different clustering methods with the Multinomial Mixture models. We experimentally evaluate the related issues, such as initialization, model estimation and generation and model selection. Besides, we propose a novel method for efficiently estimating the candidate models. Experimental results on both simulated and real data show that: (a) small run of EM (smEM) is the best choice for initialization (b) proposed hybrid model-based clustering, called EM-HAC is the best choice for candidate models estimation and (c) L-method is the best choice for model selection. As future work, we foresee the necessity to conduct further research on the model selection issue. Moreover, it is also necessary to evaluate these methods on more real-world discrete datasets obtained from a variety of different contexts.
\section*{Acknowledgments}
This work is funded by the project ImagiWeb ANR-2012- CORD-002-01.

\bibliographystyle{splncsnat}
\bibliography{ida_bib}
\end{document}